\documentclass[final]{cvpr}

\newcommand{\printfnsymbol}[1]{%
  \textsuperscript{\@fnsymbol{#1}}%
}

\usepackage{times}
\usepackage{epsfig}
\usepackage{graphicx}
\usepackage{amsmath}
\usepackage{amssymb}
\usepackage{algorithm}  

\usepackage{paralist} 
\usepackage{algorithmicx}  
\usepackage{algpseudocode}

\usepackage{algpseudocode}
\usepackage{epstopdf}
\usepackage{color}
\usepackage{cite}
\usepackage{subfigure}
\usepackage{multirow}
\usepackage{diagbox}
\usepackage{rotating}
\usepackage{booktabs}
\usepackage{colortbl}
\usepackage{overpic}
\usepackage[table]{xcolor}
\usepackage{textcomp}
\usepackage{contour}
\usepackage{textcomp}
\usepackage{adjustbox}
\usepackage{tabto,enumitem}
\usepackage{enumitem}
\usepackage{tabu}
\usepackage{siunitx,etoolbox}

\renewrobustcmd{\bfseries}{\fontseries{b}\selectfont}
\renewrobustcmd{\boldmath}{}
\newrobustcmd{\B}{\bfseries}

\makeatletter
\renewcommand{\paragraph}{%
  \@startsection{paragraph}{4}%
  {\z@}{1ex \@plus 1ex \@minus .2ex}{-1em}%
  {\normalfont\normalsize\bfseries}%
}
\makeatother

\usepackage[pagebackref=true,breaklinks=true,colorlinks,bookmarks=false]{hyperref}
\usepackage[capitalise]{cleveref}

\def\cvprPaperID{1567} 
\def\confYear{CVPR 2021}

\begin{document}

\title{Deep RGB-D Saliency Detection with Depth-Sensitive Attention\\ and Automatic Multi-Modal Fusion\vspace{-2ex}}


\author{
Peng Sun~~~~~
Wenhu Zhang~~~~~
Huanyu Wang~~~~~
Songyuan Li~~~~~
Xi Li\thanks{Corresponding Author}
\vspace{-2ex}
\\
\and
College of Computer Science \& Technology, Shanghai Institute for Advanced Study, Zhejiang University~~~~
\and
{\tt\small 
\{sunpeng1996,wenhuzhang,huanyuhello,leizungjyun,xilizju\}@zju.edu.cn}
}



\maketitle

\begin{abstract}

RGB-D salient object detection (SOD) is usually formulated as a problem of classification or regression over two modalities, \ie, RGB and depth. Hence, effective RGB-D feature modeling and multi-modal feature fusion both play a vital role in RGB-D SOD. In this paper, we propose a depth-sensitive RGB feature modeling scheme using the depth-wise geometric prior of salient objects. In principle, the feature modeling scheme is carried out in a depth-sensitive attention module, which leads to the RGB feature enhancement as well as the background distraction reduction by capturing the depth geometry prior.
 Moreover, to perform effective multi-modal feature fusion, we further present an automatic architecture search approach for RGB-D SOD, which does well in finding out a feasible architecture from our specially designed multi-modal multi-scale search space. 
 Extensive experiments on seven standard benchmarks demonstrate the effectiveness of the proposed approach against the state-of-the-art.
\end{abstract}


\section{Introduction}

Recent years have witnessed a great development of RGB-D salient object detection (SOD) 
due to its diverse applications, \eg, image retrieval~\cite{Gao20123DOR,Liu2013AMO}, video segmentation~\cite{fan2019shifting,Wang2015SaliencyawareGV}, person re-identification~\cite{zhao2013unsupervised}, visual tracking~\cite{Hong2015OnlineTB,mahadevan2009saliency}. With the multi-modal input (\ie, RGB and depth channels), 
RGB-D SOD aims to 
localize and segment
the visually salient regions in a scene, and is typically cast as an image-to-mask mapping problem within an end-to-end deep learning pipeline~\cite{peng2014rgbd,qu2017rgbd,feng2016local,fan2014salient}.

In RGB-D SOD, depth maps, which provide useful cues such as spatial structure, 3D layout, and object boundary, are important complementary information to RGB channels.
For the sake of effective learning, there are usually two key issues to solve for RGB-D SOD: 1) how to fully exploit the rich
depth geometry information for saliency analysis, and 2) how to carry out the multi-modal feature fusion effectively between RGB and depth features.
In this paper, we focus on building a depth-sensitive SOD model that is capable of learning the RGB-D feature interaction architecture automatically.

\begin{figure}[t]
   \begin{center}
   \includegraphics[width=1\linewidth]{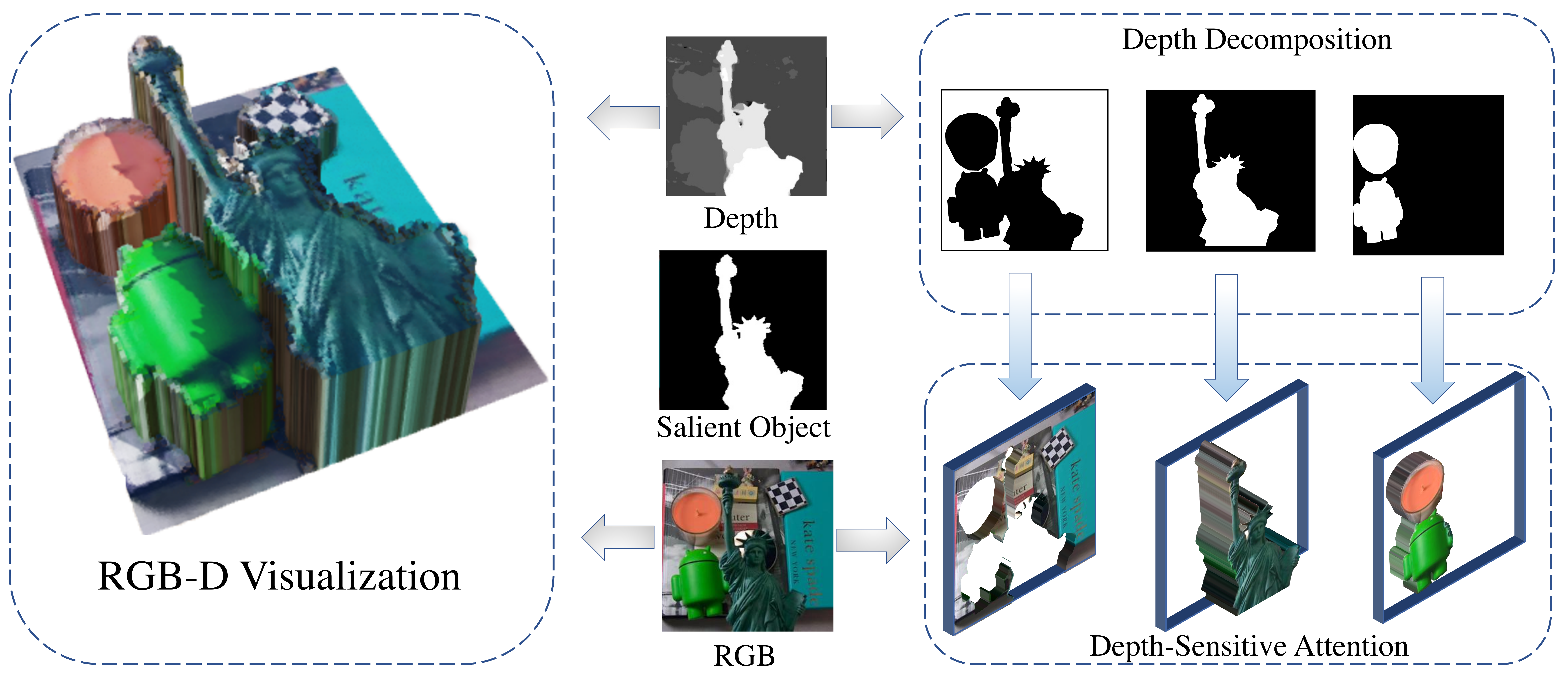}
   \end{center}
      \caption{\textbf{Left}: Salient objects are often distributed within different depth intervals. \textbf{Right}: We decompose the raw depth map into multiple regions and extract the depth-sensitive RGB features.}
   \label{fig:first_fig}
   \vspace{-15pt}
\end{figure}


In the recent literature, RGB-D SOD methods usually treat the depth channel as an auxiliary input channel, which
is directly fed into a convolutional neural network (CNN) for feature extraction~\cite{Zhang_2020_ECCV,chen2018progressively,HDFNet-ECCV2020,fan2020bbsnet,li2020}.
As a result, they are incapable of well utilizing the depth prior knowledge to capture the corresponding geometric layouts of salient objects.
As shown in \cref{fig:first_fig}, salient objects are often distributed within several particular depth intervals,
and thus can be roughly detected by regularly sliding the depth interval window. Inspired by this observation, we have an intuitive
idea that we can extract RGB features \wrt depth for effectively capturing the depth-wise geometric prior on salient objects while reducing the background distraction (\eg cluttered objects or similar texture). With this motivation, we propose to decompose the raw depth map into multiple regions, and each region contains a set of pixels from the same depth interval.
Then, we propose a depth-sensitive attention module (DSAM) to perform RGB feature extraction in different regions, thereby leading to the RGB feature enhancement with depth-wise geometric prior.

Furthermore, designing an effective feature interaction architecture between RGB and depth branches
is crucial for multi-modal feature fusion in RGB-D SOD. In general, the existing literature relies heavily on
human expertise knowledge through enormous trial and error, \eg, flow ladder module~\cite{Zhang_2020_ECCV} and
fluid pyramid integration module~\cite{zhao2019Contrast}. Moreover, the multi-source information on RGB and depth channels
is extremely heterogeneous, making the feature fusion design rather difficult and heuristic. 
Based on this observation, we leverage neural architecture search (NAS)~\cite{baker2017designing,Liu2019DARTSDA,Chen2018ReinforcedEN} 
to automatically explore an effective feature fusion module. However, simply porting existing NAS ideas from image classification/segmentation to RGB-D SOD would not suffice, as the task requires nested combinations of multi-modal multi-scale features. To this end, we construct a new search space
tailored for the multi-modal feature fusion across multiple scales for RGB-D SOD. As a result, the automatically-found feature fusion architecture equipped with the commonly used backbone VGG-19 \cite{Simonyan2015VeryDC}
achieves the state-of-the-art performance.

Our contributions can be summarized as follows:
\begin{itemize}
\item We propose a depth-sensitive attention module to explicitly eliminate the background distraction and enhance the RGB features by depth prior knowledge.


\item We design a new search space tailored for the heterogeneous feature fusion in RGB-D SOD and present the first attempt to introduce NAS for RGB-D SOD.  
\item Finally, we conduct extensive experiments on seven benchmarks, which demonstrates that our method outperforms other state-of-the-art approaches.
\end{itemize}

\section{Related Work}

\subsection{RGB-D Salient Object Detection}

Early RGB-D saliency detection methods \cite{peng2014rgbd,feng2016local,ren2015exploiting,Lang2012DepthMI} design handcrafted features, such as contrast \cite{peng2014rgbd}, shape \cite{ciptadi2013depth}, local background enclosure \cite{feng2016local} and so on. Recently, CNN-based RGB-D approaches have achieved a qualitative leap in performance due to the powerful ability of CNNs in discriminative feature representation. The existing RGB-D approaches can be roughly divided into \textbf{single-stream} models \cite{peng2014rgbd,DANet,Song2017DepthAwareSO,Liu2019SalientOD,Zhu2019PDNetPG,Shigematsu2017LearningRS} and \textbf{multi-stream} models \cite{chenhaommf,Zhang_2020_ECCV,chen2018progressively,HDFNet-ECCV2020,fan2020bbsnet,Chen2019ThreeStreamAN,Chen2019MultimodalFN,Chen2020DiscriminativeCT}. 
The single-stream architecture adopts a straightforward way to fuse RGB images and depth cues. For example, Peng \etal \cite{peng2014rgbd} directly concatenate RGB-D pairs as 4-channel inputs to predict saliency maps. DANet \cite{DANet} uses a single-stream network with the depth-enhanced dual attention for salient object detection.
For the multi-stream models, the frameworks employ two parallel networks to extract RGB and depth features respectively, and then fuse the multi-modal features with various dazzling strategies. For example, Chen \etal \cite{chenhaommf} design a multi-branch network
to fuse the deep and shallow cross-modal complements in
separate paths, and then propose to use residual connections and complementarity-aware supervisions to explicitly expose cross-modal complements in~\cite{chen2018progressively}. Lately, Zhang~\cite{Zhang_2020_ECCV} proposes an asymmetric two-stream architecture, and designs a flow ladder module for the RGB stream and a depth attention module for the depth stream.

Although these methods have achieved huge success, depth cues are only direct as the input of the feature extractor. In this paper, motivated by our observation, we further exploit the depth information, which contains abundant geometric prior knowledge. Then, we utilize the depth cues to explicitly eliminate the background distraction and propose an effective depth-sensitive attention module for RGB-D salient object detection.

\begin{figure*}
\begin{center}
\includegraphics[width=6.9in]{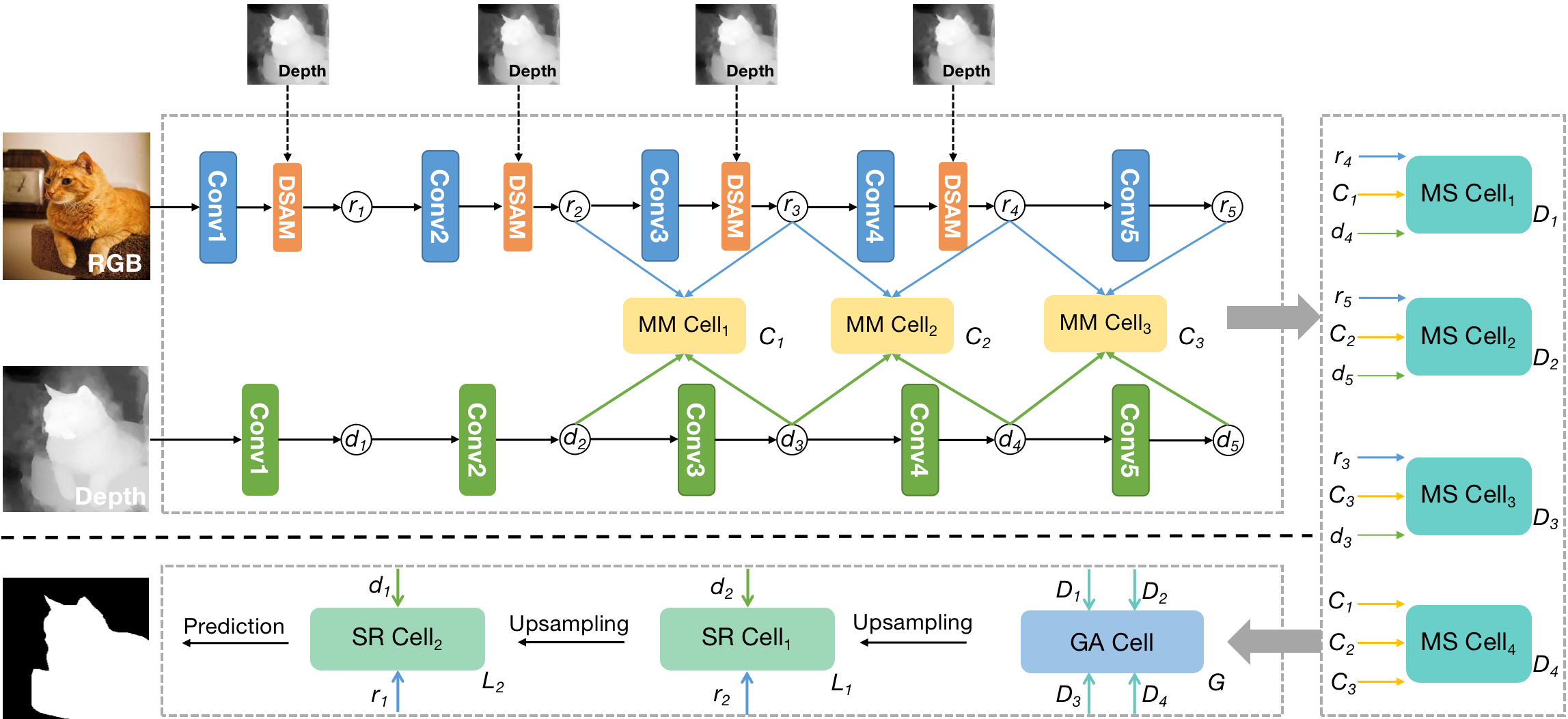}
\end{center}
   \caption{Illustration of our proposed framework. The whole network consists of an RGB branch, a depth branch, and a specially-designed module. The RGB branch is equipped with the proposed depth-sensitive attention modules (DSAMs). $r_i$, $d_i$, $C_i$, $D_i$, $G$, and $L_i$ represent the output features, and thin arrows represent the feature flows. \textbf{Best viewed in color.}}
\label{fig:main_fig}
\vspace{-12pt}
\end{figure*}

\subsection{Neural Architecture Search}

Neural architecture search (NAS) aims at automating the network architecture design process.
Early NAS works are based on either reinforcement learning \cite{baker2017designing, DBLP:conf/iclr/ZophL17} or evolutionary algorithms \cite{Real2019RegularizedEF, Chen2018ReinforcedEN}. Despite achieving satisfactory performance, they have consumed hundreds of GPU days. 
Recently, one-shot methods \cite{Bender2018UnderstandingAS,Brock2018SMASHOM} have greatly solved the time-consuming problem by training a parent network from which each sub-network can inherit the weights. DARTs \cite{Liu2019DARTSDA} is the pioneering work for gradient-based NAS, 
which uses gradients to efficiently optimize the search space. 
After that, NAS has been widely applied to many computer vision tasks, such as object detection \cite{Ghiasi2019NASFPNLS,Xu2019AutoFPNAN}, semantic segmentation \cite{autodeeplab2019,Lin2020GraphGuidedAS},  and so on.

However, in RGB-D salient object detection, the multi-modal feature fusion architectures are still designed by hand. Although there are several NAS works \cite{Yu2020DeepMN,PrezRa2019MFASMF} for multi-modal fusion, their design purpose is especially for the visual question answering task \cite{Yu2020DeepMN} or image-audio fusion task \cite{PrezRa2019MFASMF}. As far as we know, our work is the first attempt to utilize the NAS algorithms to tackle the multi-modal multi-scale feature fusion problem for RGB-D SOD.


\section{Method}

In this section, we illustrate the proposed \textbf{d}epth-\textbf{s}ensitive \textbf{a}ttention and \textbf{a}utomatic multi-modal \textbf{f}usion (DSA$^2$F) framework in detail.
First, we briefly introduce an overview of the proposed framework.
Then, we describe the proposed depth-sensitive attention.
Next, we elaborate on the task-specific module for the automatic multi-modal multi-scale feature fusion. Finally, we illustrate the whole optimization strategy.

\subsection{Overview}

In DSA$^2$F, the whole network consists of an RGB branch, a depth branch, and a specially-designed fusion module, as shown in \cref{fig:main_fig}. The RGB branch is based on VGG-19~\cite{Simonyan2015VeryDC}, and the depth branch is a lightweight depth network 
to obtain the depth features of different scales.

We plug in a depth-sensitive attention module (DSAM) following each down-sampling layer in the RGB branch. Each DSAM utilizes a raw depth map to enhance the RGB features. Specifically, we decompose the raw depth map into multiple regions. Each region, which contains the pixel values from the same depth distribution mode,
is considered as a spatial attention map to extract the corresponding RGB features.




To fuse the enhanced RGB features and the depth features automatically, 
we propose a multi-modal multi-scale feature fusion module. 
In the RGB-D SOD literature~\cite{Zhang_2020_ECCV,li2020,piao2019depth,DANet,fan2020bbsnet,Li_2020_CMWNet,Zhang2020SelectSA}, three consistent principles are noticeable: 
\begin{inparaenum}[1)]
   \label{principle}
   \item The features from different modalities of the same scale are always fused, while 
   features in different scales are selectively fused.
   \item Low-level features are always combined with high-level features before the final prediction, as low-level features are rich in spatial details but lack semantic information and vice versa.
   \item Attention mechanism is necessary when performing the feature fusion of different modalities. 
\end{inparaenum}
With these common practices, we design a new search space adapted to the multi-modal multi-scale fusion, which contains four different architectures \ie, the \textbf{m}ulti-\textbf{m}odal fusion (MM), \textbf{m}ulti-\textbf{s}cale fusion (MS), \textbf{g}lobal context \textbf{a}ggregation (GA) and \textbf{s}patial information \textbf{r}estoration (SR) cells.


\subsection{Depth-Sensitive Attention}



\begin{figure*}[t]
   \centering
   \vspace{-5pt}
	\subfigure[Depth decomposition]{
		\centering
		\includegraphics[width=0.30\linewidth]{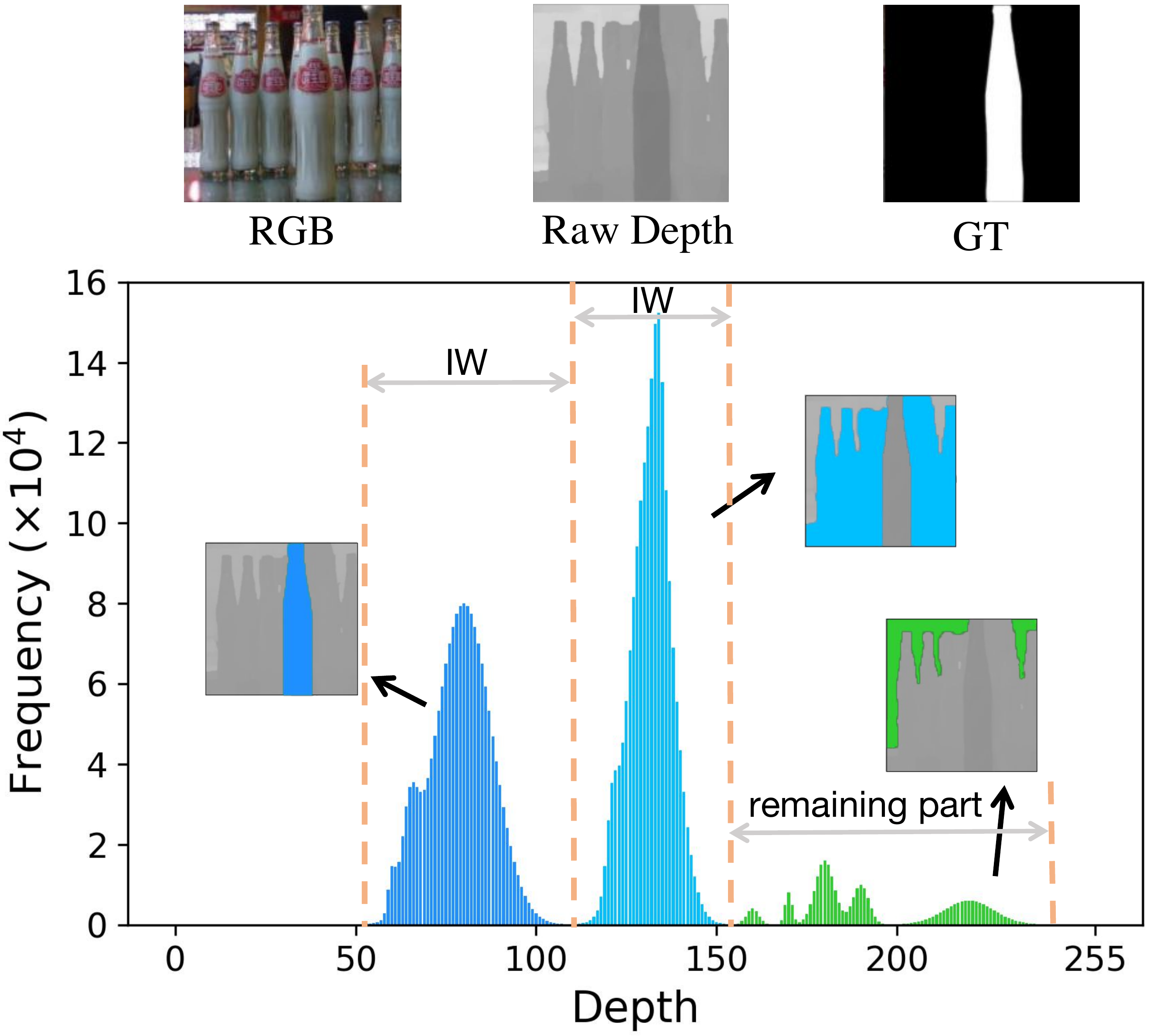}
		\label{DSAM.figa}
	}
	\subfigure[Depth-sensitive attention module]{
		\centering
		\includegraphics[width=0.62\linewidth]{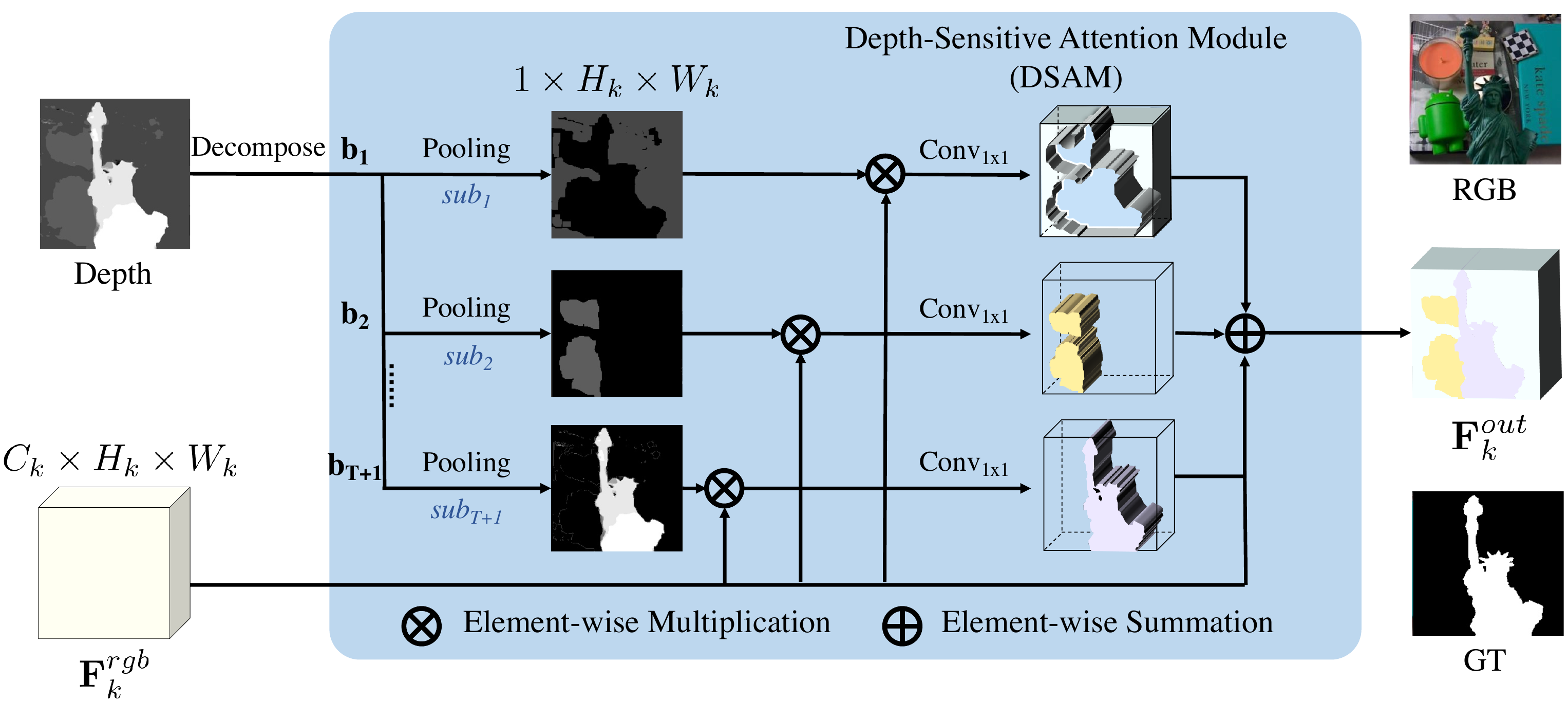}
		\label{DSAM.figb}
	}
   \caption{
   (a) The depth decomposition process. `IW' represents the depth interval window here. 
   (b) The detailed depiction of the proposed depth-sensitive attention module. \textbf{Best viewed in color.}
   }
   \vspace{-15pt}
	\label{fig.DSAM}
\end{figure*}

We propose a depth-sensitive RGB feature modeling scheme, including the depth decomposition and the depth-sensitive attention module. 
The raw depth map is decomposed into $T+1$ regions with the following steps.
First, we quantize the raw depth map into the depth histogram, and 
choose the 
$T$ largest depth distribution modes (corresponding to the $T$ depth interval windows) of the depth histogram.
Then, using these depth interval windows, the raw depth map can be decomposed into $T$ regions, and the remaining part of the histogram naturally forms the last region, as shown in \cref{DSAM.figa}.
Finally, each region is normalized into [0,1] as a spatial attention mask for the subsequent process.
 

After obtaining these attention masks, we describe the depth-sensitive attention module in detail.
In DSAM, the obtained attention masks give rise to $T+1$ sub-branches in the RGB branch, as shown in \cref{DSAM.figb}.
Formally, let $\mathbf{F}_{k}^{rgb}$ $\in$ $\mathbb{R}^{C_k\times H_k\times W_k}$ be the RGB feature maps in the $k$-th stage of the RGB branch, where $C_k$, $H_k$, and $W_k$ represent the number of channels, the height, and width, respectively. 
Denote $b_t$ as the $t$-th attention mask obtained in the above depth decomposition process. 
We utilize the max-pooling operation to align the masks to the size of $\mathbf{F}_{k}^{rgb}$ as
\begin{equation}
 p_t = \mathrm{MaxPool}(b_{t}),
\end{equation}
where $p_t \in \mathbb{R}^{H_k\times W_k}$.
Next, we utilize the resized masks, $\{p_1,p_2, \cdots, p_{T+1}\}$, to extract the depth-sensitive features in $T+1$ parallel sub-branches. 
Specifically, we multiply each mask $p_{t}$ with each channel of RGB features $\mathbf{F}_{k}^{rgb}$ and use a 1 $\times$ 1 convolution layer in $t$-th sub-branch as a transition layer, to refine the RGB features from various depth intervals. 
After that, we aggregate all the depth-sensitive features from $T+1$ sub-branches by an element-wise summation operation, 

\begin{equation}
   F_{k}^{enh} = \sum_{t=0}^{T} \mathrm{Conv}_{1 \times 1}(p_t \otimes F_{k}^{rgb})
   \end{equation}
where $F_{k}^{enh}$ is the enhanced RGB features and $\otimes$ indicates the element-wise multiplication.
Finally, we introduce a residual connection and get the final output features, 
\begin{equation}
   r_{k} = F_{k}^{enh}+ F_{k}^{rgb}.
\end{equation}


In this way, DSAM not only provides depth-wise geometric prior knowledge for RGB features, but also eliminates the intractable background distraction (\eg cluttered objects or similar texture).
Furthermore, the ablation experiments in \cref{para:ablation-dsam} also verify the effectiveness of our DSAM.

\subsection{Auto Multi-Modal Multi-Scale Feature Fusion}

We propose an automatic multi-modal multi-scale fusion module for RGB-D SOD.
First, we describe the designed four types of cells, \ie, MM, MS, GA, SR cells, and they build up the entire task-specific search space.
Then, we elaborate on the search space of the fusion module, in which the cells of four types cooperate in a sequential pipeline.
Finally, we describe the internal structure of each cell.

\paragraph{Cell types.}
For RGB-D SOD,
we design four types of cells and each cell is a searchable unit in NAS. 
First, we use \textit{MM cells} to directly perform multi-modal feature fusion between RGB and depth branches.
Second, we use \textit{MS cells} for the dense multi-scale feature fusion. 
Third, we utilize \textit{GA cell} to aggregate seamlessly the outputs of the MS cells for capturing the global context. 
Finally, we introduce \textit{SR cells} to combine the low-level and high-level features to remedy the spatial detail loss caused by downsampling. 
The whole execution process in the proposed search space is detailed as follows.

\paragraph{Search space.} By the searchable fusion module, we fuse the RGB features $\{r_1, r_2, \cdots, r_5\}$
with depth features 
$\{d_1, d_2, \cdots, d_5\}$
as shown in \cref{fig:main_fig}. 
Specifically, first, we take the adjacent features from both branches as the input of MM cells, to obtain the multi-modal features:
\begin{equation}
   C_{n} =\mathrm{MM_{n}} (r_{n+1}, r_{n+2}, d_{n+1}, d_{n+2}), n \in \{1,2,3\},
\end{equation}
where $C_{n}$ is the output of the $n$-th CM cell.

Next, we carry out further dense feature fusion through MS cells. There are two kinds of multi-scale fusion, \ie, to fuse each multi-modal feature with original features in different scales by three MS cells, and to fuse all the derived multi-modal multi-scale features by another MS cell. The process can be represented as:
\begin{equation}
   \begin{aligned}
   D_{m} =
   \left\{
   \begin{array}{lr}
   \mathrm{MS_{m}}(r_4, C_1, d_4),&m = 1,  \\
   \mathrm{MS_{m}}(r_5, C_2, d_5),&m = 2,  \\
   \mathrm{MS_{m}}(r_3, C_3, d_3),&m = 3,  \\
   \mathrm{MS_{m}}(C_1, C_2, C_3),&m=4,
   \end{array}
   \right.
   \end{aligned}
\end{equation}
where $m$ is the index of the MS cells. 



After that, a GA cell is introduced to seamlessly integrate the outputs of the above four MS cells for global context aggregation, which is calculated by:
\begin{equation}
   G =\mathrm{GA}(\{D_m\}), m \in \{1,2,3,4 \}.
\end{equation}

Finally, to compensate the loss of spatial detail caused by downsampling, we use two sequential SR cells to fuse the high-level features $G$ and the low-level features (\ie $r_1$, $d_1$ or $r_2$, $d_2$) as:
\begin{equation}
   \begin{aligned}
   L_1 =\mathrm{SR}_{1} (\mathrm{ \sigma }(G), d_2, r_2),\\
   L_2 =\mathrm{SR}_{2} (\sigma(L_1), d_1, r_1),
   \end{aligned}
\end{equation}
where $\sigma$ indicates the upsampling function. In the end, a simple decoder is adopted for supervision. The decoder contains two bilinear upsampling functions, each of which is followed by three convolutional layers.



\paragraph{Cell structure.}


 Each aforementioned cell can be formulated by a unified structure, which is a directed acyclic graph (DAG) consisting of an ordered sequence of $N$ nodes, denoted by $\mathcal{N}$ = $\{x^{(1)},...,x^{(N)}\}$. Each node $x^{(i)}$ is a latent representation (\ie feature map), and each directed edge $\left(i, j \right)$ is associated with some candidate operations $o^{(i, j)} \in \mathcal{O} $  (\eg conv, pooling), representing all possible transformations from $x^{(i)}$ to $x^{(j)}$. 
Each intermediate node $x^{(j)}$ is computed based on all of its predecessors:

\begin{equation}
x^{(j)} = \sum\nolimits_{i < j} o^{(i, j)} \big( x^{(i)} \big).
\end{equation}
 
 To make the search space continuous, we relax the categorical choice of a particular operation to a softmax over all possible operations \cite{Liu2019DARTSDA}:
 
\begin{equation}
\widetilde{o}^{(i, j)}(x)  =  \sum\nolimits_{o \in \mathcal{O}}  \mathrm{Softmax}(\alpha_{o}^{(i,j)})     o(x),
\end{equation}
where $o(\cdot)$ is an operation in the operation set $\mathcal{O}$, and $\alpha_{o}^{(i,j)}$ is the learnable architecture parameter of the operation selection for edge $\left(i, j \right)$. Thus, each cell architecture is denoted by \{$\alpha^{(i,j)}$\}. The whole searchable fusion module can be represented as 
$\alpha$ = $\{ \alpha_{mm}, \alpha_{ms}, \alpha_{ga}, \alpha_{sr} \}$.
Cells of the same type share the same architecture parameters, but with different weights.
After the searching phase, an optimal operation can be determined by replacing each mixed operation $\widetilde{o}^{(i, j)}$ with the most likely operation (\ie $\text{argmax}_{o \in \mathcal{O}} \alpha_{o}^{(i,j)})$. 

\paragraph{Discussion.} Let us retrospect the three consistent principles in the RGB-D literature, as discussed in \cref{principle}. 
Our task-specific search space is general enough to cover the above mentioned common practices. 
To be specific, the design philosophy for MM and MS cells meets the requirement of multi-modal feature fusion in not only the same scale but also different scales. 
Then, the GA cell introduces the low-level spatial information to the high-level features. 
Moreover, we add the spatial and channel attention operations into the candidate operation set $\mathcal{O}$ to explore the collocation of attentions, and detailed analysis can be found in \cref{para:ablation-atten}.

\subsection{Optimization}

The optimization of our framework consists of two stages. First, we search the multi-modal fusion module. Then, we optimize the whole network.

\paragraph{Multi-modal fusion module search.}

During the search progress, we hold out half of the original training data as the validation set.
We use the bi-level optimization \cite{Anandalingam1992HierarchicalOA,Colson2007AnOO} to jointly optimize architecture parameter $\alpha$ and network weights $w$:

\begin{equation}
	\begin{array}{cl}
		\min\limits_{\alpha} & \mathcal{L}_{v a l}\left(w^{*}(\alpha), \alpha\right) \\
		\text { s.t. } & w^{*}(\alpha)=\operatorname{argmin}_{w} \mathcal{L}_{train}(w, \alpha),
   \end{array}
   \label{equ:nas}
\end{equation}
where $  \mathcal{L}_{v a l}$ and $\mathcal{L}_{train}$ denote validation loss and training loss (both are the cross-entropy loss), respectively.
Then the fusion module is obtained by the discrete $\alpha$ by \cref{equ:nas}.

\paragraph{The whole network optimization.} With the obtained fusion module, the whole network is optimized on the whole training data by the standard cross-entropy loss for the saliency detection.

\begin{equation}
	\begin{array}{cl}
      w^{*}= \min\limits_{w} & \mathcal{L}_{train}\left(w, \alpha\right).
	\end{array}
\end{equation}

\section{Experiments}

In this section, we conduct extensive experiments to verify the effectiveness of our method. Firstly, we compare our DSA$^2$F with other state-of-the-art methods on seven standard benchmarks. Secondly, we perform a series of ablation studies to evaluate each component of our framework.

\subsection{Datasets and Evaluation Metrics}

\paragraph{Datasets.} We perform our experiments on seven widely used RGB-D datasets for fair comparisons, including DUT-RGBD \cite{piao2019depth}, NJUD \cite{ju2014depth}, NLPR \cite{peng2014rgbd}, SSD \cite{zhu2017three}, STEREO \cite{niu2012leveraging}, LFSD \cite{li2014saliency} and RGBD135 \cite{cheng2014depth}.
To guarantee fair comparisons, we choose the same 800 samples from DUT-RGBD, 700 samples from NLPR and 1485 samples from NJUD as ATSA~\cite{Zhang_2020_ECCV} to train our model. The remaining images and other datasets are for testing to comprehensively verify the generalization ability of saliency models.

\paragraph{Evaluation metrics.} To comprehensively and fairly evaluate various methods, we employ four widely used metrics, including mean F-measure ($\mathcal{F}_{\beta}$) \cite{achanta2009frequency}, mean absolute error ($\mathcal{M}$) \cite{borji2015salient}, S-measure ($\mathcal{S}_{\lambda}$) \cite{fan2017structure}, E-measure ($\mathcal{E}_{\xi}$) \cite{fan2018enhanced}. Specifically, the F-measure can evaluate the overall  performance based on the region similarity. The $\mathcal{M}$ measures the average of the per-pixel absolute difference between the saliency maps and the ground truth. The S-measure that is recently proposed can evaluate the structural similarities. The E-measure can jointly utilize image-level statistics and local pixel-level statistics for evaluating the binary saliency map.

\begin{table*}[t!]
  \centering
  \small
  \renewcommand{\arraystretch}{1.7}
  \renewcommand{\tabcolsep}{0.8mm}
  \caption{\small
  \textbf{Quantitative results of 18 state-of-the-art methods on seven datasets}:
  \textit{DUT-RGBD}~\cite{piao2019depth},
  \textit{NJUD}~\cite{ju2014depth},
  \textit{NLPR}~\cite{peng2014rgbd},
  \textit{SSD}~\cite{zhu2017three},
  \textit{STEREO}~\cite{niu2012leveraging},
  \textit{LFSD}~\cite{li2014saliency},
  and \textit{RGBD135}~\cite{cheng2014depth}.
   $\uparrow$ and $\downarrow$ stand for larger and smaller is better, respectively. The best results are marked in \textcolor{red}{\underline{red}}.
    The column `Pub.' denotes the publication of each method.
  }\label{tab:ModelScore}
\resizebox{1\textwidth}{!}{
\begin{tabular}{l|c|cccc|cccc|cccc|cccc|cccc|cccc|cccc}
\midrule[1.5pt]    
 \multirow{2}{*}{\normalsize{Method}} 
 & \multirow{2}{*}{\normalsize{Pub.}}
 & \multicolumn{4}{c|}{DUT-RGBD} 
 & \multicolumn{4}{c|}{NJUD} 
 & \multicolumn{4}{c|}{NLPR} 
 & \multicolumn{4}{c|}{SSD} 
 & \multicolumn{4}{c|}{STEREO}
 & \multicolumn{4}{c|}{LFSD} 
 & \multicolumn{4}{c|}{RGBD135}
 \\
 \cmidrule(l){3-6} \cmidrule(l){7-10} \cmidrule(l){11-14} \cmidrule(l){15-18} \cmidrule(l){19-22} \cmidrule(l){23-26} \cmidrule(l){27-30}
          &   & $E_{\gamma}\uparrow$ & $\mathcal{S}_{\lambda}\uparrow$ & $\mathcal{F}_{\beta}\uparrow$  & $ \mathcal{M}\downarrow$
   	   & $E_{\gamma}\uparrow$ & $\mathcal{S}_{\lambda}\uparrow$ & $\mathcal{F}_{\beta}\uparrow$  & $ \mathcal{M}\downarrow$
	    & $E_{\gamma}\uparrow$ & $\mathcal{S}_{\lambda}\uparrow$ & $\mathcal{F}_{\beta}\uparrow$  & $ \mathcal{M}\downarrow$
	     & $E_{\gamma}\uparrow$ & $\mathcal{S}_{\lambda}\uparrow$ & $\mathcal{F}_{\beta}\uparrow$  & $ \mathcal{M}\downarrow$
	     & $E_{\gamma}\uparrow$ & $\mathcal{S}_{\lambda}\uparrow$ & $\mathcal{F}_{\beta}\uparrow$  & $ \mathcal{M}\downarrow$
	     & $E_{\gamma}\uparrow$ & $\mathcal{S}_{\lambda}\uparrow$ & $\mathcal{F}_{\beta}\uparrow$  & $ \mathcal{M}\downarrow$
	     & $E_{\gamma}\uparrow$ & $\mathcal{S}_{\lambda}\uparrow$ & $\mathcal{F}_{\beta}\uparrow$  & $ \mathcal{M}\downarrow$
         \\

\midrule[1pt]

\textbf{TANet}~\cite{Chen2019ThreeStreamAN} & TIP19 &
.866 & .808 & .799 & .093 &  
.893 & .878 & .844 & .061 &  
.916 & .886 & .795 & .041 &   
.879 & .839 & .767 & .063 &   
.911 & .877 & .849 & .060 &  
.845 & .801 & .794 & .111 &  
.916 & .858 & .782 & .045 \\   

\textbf{CPFP}~\cite{zhao2019Contrast} & CVPR19
&.814 & .749 & .736 & .099 & 
.906 & .878 & .877 & .053 & 
.924 & .888 & .822 & .036 & 
.832 & .807 & .725 & .082 & 
.897 & .871 & .827 & .054 &
.867 & .828 & .813 & .088 &
.927 & .874 & .819 & .037 \\

\textbf{DMRA}~\cite{piao2019depth} & ICCV19
&.927 & .888 & .883 & .048 & 
.908 & .886 & .872 & .051 & 
.942 & .899 & .855 & .031 & 
.892 & .857 & .821 & .058 & 
.920 & .886 & .868 & .047 &
.899 & .847 & .849 & .075 &
.945 & .901 & .857 & .029  \\

\textbf{A2dele}~\cite{Piao2020A2deleAA} & CVPR20 &
- & - & .892 & .042 &  
- & - & .874 & .051 &  
- & - & .878 & .028 &   
- & - & - & - &   
- & -& .884 & .043 &  
- & - & - & - &  
- & - & .865 & .028 \\   

\textbf{S2MA}~\cite{Liu2020LearningSS} & CVPR20 &
.921 & .903 & .886 & .043 &  
\textcolor{red}{\underline{.930}} & .894 & - & .053 &  
.937 & .915 & .847 & .030 &   
.891 & .868 & .818 & .053 &   
.907 & .890 & .855 & .051 &  
.863 & .837 & .803 & .095 &  
\textcolor{red}{\underline{.971}}
 & 
\textcolor{red}{\underline{.941}}
  & .893 & .021 \\   

\textbf{UC-Net} \cite{Zhang2020UCNetUI} & CVPR20 &
- & - & - & - &  
\textcolor{red}{\underline{.930}} & .897 & .886 & .043 &  
.951 & .920 & .886 & .025 &   
- & - & - & - &   
.922 & .903 & .885 & .039 &  
.897 & .865 & .859 & .066 &  
.967 & .934  & .905 & 
\textcolor{red}{\underline{.019}}\\   

\textbf{JL-DCF} \cite{Fu2020JLDCF} & CVPR20 &
.931 & .906 & .882 & .043 &  
- & .903 & - & .043 &  
.952
 & 
 \textcolor{red}{\underline{.925}}
& .875 & 
\textcolor{red}{\underline{.022}} &   
- & - & - & - &   
.919 & .903 & .869 & .040 &  
.882 & .862 & .854 & .070 &  
.965 & .929 & .885 & .022 \\

\textbf{SSF}~\cite{Zhang2020SelectSA} & CVPR20 &
- & .915 & .915 & .033 &  
- & .899 & .886 & .043 &  
- & .914 & .875 & .026 &   
- & - & - & - &   
- & .893 & .880 & .044 &  
- & .859 & .867 & .066 &  
- & .905 & .876 & .025 \\   

\textbf{D$^3$Net} \cite{Fan2020RethinkingRS} & TNNLS20 &
- & - & - & - &  
.913 & .900 & .863 & .047 &  
.943 & .912 & .857 & .030 &   
.897 & .857 & .802 & .058 &   
.920 & .899 & .859 & .046 &  
.853 & .825 & .789 & .095 &  
.951 & .900 & .859 & .030 \\   

\textbf{CoNet} \cite{Wei_2020_ECCV} & ECCV20 &
.941 & .918 & .908 & .034 &  
.912 & .895 & .872 & .047 &  
.934 & .908 & .848 & .031 &   
.896 & .853 & .806 & .059 &   
.924 & 
.908
 & 
 .885
 & .040 &  
.896 & .862 & .848 & .071 &  
.945 & .911 & .862 & .027 \\   

\textbf{ATSA}~\cite{Zhang_2020_ECCV} & ECCV20 
&
.948 & .918 & .920 & .032
& .921& .901 & .893 & .040 &
.945 & .907 & .876 & .028 & 
.901 & .860 & .827 & .050 & 
.921 & .897 & .884 & .039 &
.905 & .865 & .862 & .064 &
.952 & .907 & .885 & .024  \\

\textbf{CMMS}~\cite{li2020} & ECCV20 &
.940 & .913 &.906&.037&  
.914&.900 &.886&.044 & 
.945&.915 & .869&.027&   
.911&.874 &.842 & 
.046 & 
.922 &.895 &.879 &.043 & 
.891 &.849 &.869&.073 & 
- & - & - & - \\   

\textbf{BBS-Net}~\cite{fan2020bbsnet} & ECCV20 &
- & - & - & - &  
.918 & 
\textcolor{red}{\underline{.917}}  
& 
.899 & 
.037 &  
\textcolor{red}{\underline{.954}}
 & 
.924 & .880 & .025 &   
.890 & .855 & .806 & .056 &   
.920 & .901 & .876 & .043 &  
.889 & .852 & .843 & .074 &  
.951 & .918 & .871 & .025 \\   

\textbf{PGAR} \cite{chen2020eccv} & ECCV20 &
.944 & 
.919
 & .913 & .035 &  
.916 & .909 & .893 & .042 &  
\textcolor{red}{\underline{.954}}
 & .930 & .883 & .025 &   
- & - & - & - &   
.919 & 
\textcolor{red}{\underline{.913}}
 & .880 & .041 &  
.889 & .853 & .852 & .074 & 
.940 & .916 & .869 & .025 \\   

\textbf{CAS-GNN}~\cite{Luo2020CascadeGN} & ECCV20
&
.932 & .891 & .912 & .043 &  
.922 & 
.911
 & .882 & 
 \textcolor{red}{\underline{.036}} 
 &  
.951 & .919 & 
.888
 & .025 &   
 \textcolor{red}{\underline{.915}}
  & .872 & 
.840
&.047 &   
.929 & .899 & .876 & 
.039 &  
.877 & .846 & .832 & .074 & 
.943 & .898 & .885 & .026 \\ 

\textbf{CMWNet} \cite{Li_2020_CMWNet} & ECCV20 &
.916 & .887 & .865 & .056 &  
.911 & .903 & .879 & .046 &  
.939 & .917 & .857 & .029 &   
.900 & 
.875
& .819 & .051 &   
.917 & .905 & .869 & .043 &  
.890 & 
.876 & .870
 & .066 &  
.967 & .937 & .889 & .021 \\   

\textbf{DANet}~\cite{DANet} & ECCV20 &
.934 & .899 & .883 & .043 &  
.922 & .899 & .871 & .045 &  
.949 & .915 & .870 & .028 &   
.911 & .864 & .827 & .050 &   
.921 & .901 & .868 & .043 &  
.874 & .849 & .822 & .079 &  
.967
 & .924 & .887 & .023 \\   

\textbf{HDFNet} \cite{HDFNet-ECCV2020} & ECCV20 &
 .938 & .905 & .865 & .040 &  
.920 & .885 & .847 & .051 &  
.942 & .898 & .839 & .031 &   
.913
 & .866 & .808 & .048 &   
 \textcolor{red}{\underline{.937}}
  & .906 & .863 & 
.039&  
.899 & .847 & .811  & .076 &  
.944 & .899 & .843 & .030 \\   

\midrule[1pt]

\textbf{Ours} & -
&
\textcolor{red}{\textbf{\underline{.950}}} & 
\textcolor{red}{\textbf{\underline{.921}}} & 
\textcolor{red}{\textbf{\underline{.926}}} &
\textcolor{red}{\textbf{\underline{.030}}} & 

.923 & .903 & 
\textcolor{red}{\textbf{\underline{.901}}} &
 .039 &  


.950 & .918 & 
\textcolor{red}{\textbf{\underline{.897}}} &
.024 &   

.904 & 
\textcolor{red}{\textbf{\underline{.876}}} & 
\textcolor{red}{\textbf{\underline{.852}}} &
\textcolor{red}{\textbf{\underline{.045}}} &

.933 & .904 & 
\textcolor{red}{\textbf{\underline{.898}}} & 
\textcolor{red}{\textbf{\underline{.036}}} &  

\textcolor{red}{\textbf{\underline{.923}}} & 
\textcolor{red}{\textbf{\underline{.882}}} & 
\textcolor{red}{\textbf{\underline{.882}}} & 
\textcolor{red}{\textbf{\underline{.054}}} & 

.962 & .920 & 
\textcolor{red}{\textbf{\underline{.896}}} & .021

\\ 
                 
\toprule[1.5pt]
\end{tabular}
}
 \vspace{-10pt}
\end{table*}

\subsection{Implementation Details}

Our method is implemented with PyTorch toolbox \cite{Paszke2017AutomaticDI}. For the depth branch, we use the DepthNet\cite{Zhang_2020_ECCV} which is a lightweight network compared with VGG-19. For the depth-sensitive attention module, the number of depth decomposition regions is 3. 
In the search process, the node numbers of the MM, MS, GA, SR cells are 8, 8, 8, 4, respectively. 
For the candidate operation set $\mathcal{O}$, we collect the $\mathcal{O}$ as follows: max pooling, skip connection, 3 $\times$ 3 conv, 1 $\times$ 1 conv, 3 $\times$ 3 separable conv, 3 $\times$ 3 dilated conv (dilation=2), 3 $\times$ 3 spatial attention and 1 $\times$ 1 channel attention. 
For the training hyper-parameters, the batch size is set to 8. The architecture parameters $\alpha$ are optimized by Adam, with an initial learning rate 3e-4, a $\beta$ = (0.5, 0.999) and a weight decay 1e-3. The network parameters are optimized using SGD with an initial learning rate of 0.025, a momentum of 0.9 and a weight decay of 3e-4. The search process contains 50 epochs and takes approximately 20 hours on 4 GTX 1080Ti GPUs. 

After searching, the network is trained on a GTX 1080Ti GPU, and the input images are uniformly resized to 256 $\times$ 256. The momentum, weight decay and learning rate of our network are set as 0.9, 5e-4 and 1e-10, respectively. The network converges after 60 epochs with mini-batch size 2. To reduce overfitting, we augment the training set by randomly flipping, cropping and rotating the training images.

\subsection{Comparison with State-of-the-art}

We compare our DSA$^2$F with 18 other state-of-the-art methods on seven widely-used benchmarks, and for a fair comparison, we recalculate the mean F-measure of other methods according to their provided saliency maps if they report the max F-measure in the paper. 

\paragraph{Quantitative comparison.}~\cref{tab:ModelScore} shows the quantitative comparison in terms of four evaluation metrics on seven datasets. 
All results in the table are quoted or tested by VGG-19~\cite{Simonyan2015VeryDC} backbone for a fair comparison.
It can be seen that DSA$^{2}$F significantly outperforms the competing methods across all the datasets in most metrics. Especially, DSA$^{2}$F outperforms all other methods by a dramatic margin on the LFSD and DUT-RGBD dataset, which are considered as more challenging datasets due
to the large number of complex scenes like similar foreground and background,
low-contrast and transparent object. Moreover, DSA$^{2}$F consistently surpasses all other state-of-the-art methods in seven datasets in terms of the overall performance metric (\ie $\mathcal{F}_{\beta}$).

\begin{figure*}[t]
   \centering
   \vspace{-0.2cm}
   \includegraphics[width=6.5in]{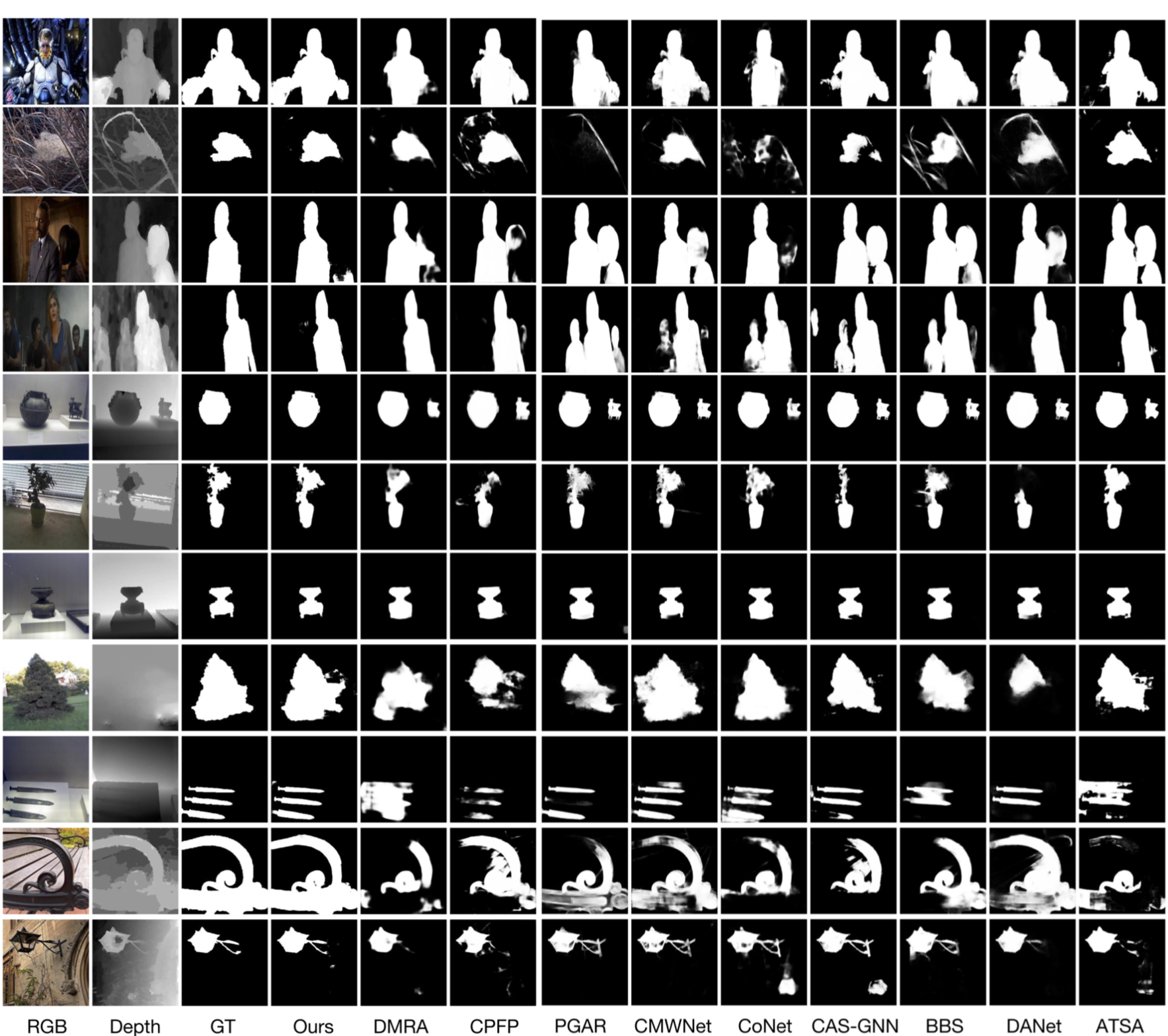}
   \caption{Qualitative comparison of the state-of-the-art RGB-D SOD methods and our approach. Obviously, saliency maps produced
   by our model are clearer and more accurate than others in various challenging scenarios.}
   \label{fig:show_citys}
   \vspace{-14pt}
   \end{figure*}

\paragraph{Qualitative comparisons.} To further illustrate the superior performance of our method, Fig \ref{fig:show_citys} shows some visual results of the proposed method and other state-of-the-art methods. From those results, we can observe that our method is able to accurately segment salient objects under various challenging scenarios, including images with low contrast foreground and background (1$^{st}$ and 2$^{nd}$ rows), cluttered distraction objects (3$^{rd}$, 4$^{th}$ and 5$^{th}$ rows), blurry depth (8$^{th}$ and 9$^{th}$ rows), and fine structures (10$^{th}$ and 11$^{th}$ rows). These results further demonstrate our approach could eliminate the background distraction obviously in utilizing the depth prior knowledge. Moreover, the object boundaries (6$^{th}$ and 7$^{th}$ rows) of our results are more clear and sharper than others, which preserves more details.

\subsection{Ablation Analysis}

In this section, we perform a series of ablation studies to further investigate the relative importance and specific contribution of each component in the proposed framework.

\paragraph{Effectiveness of depth-sensitive attention module.}\label{para:ablation-dsam}

\begin{table}[t]
 \caption{\small Ablation study for DSAM on three widely-used datasets.}
 \label{tab:agg}
 \centering
 \footnotesize
 \begin{tabu} to 0.99\columnwidth{X[0.5,c] | X[4,c]| X[c] X[c] | X[c] X[c] | X[c] X[c]}
  \toprule
  \multicolumn{1}{c|}{\multirow{2}{*}{{\#}}}&
  \multicolumn{1}{c|}{\multirow{2}{*}{Settings}} & \multicolumn{2}{c|}{DUT-RGBD~} &
  \multicolumn{2}{c|}{NLPR~}& 
  \multicolumn{2}{c}{SSD~} 
  \\
  && 
  $\mathcal{F}_{\beta}\uparrow$  & $ \mathcal{M}\downarrow$&
  $\mathcal{F}_{\beta}\uparrow$  & $ \mathcal{M}\downarrow$&
  $\mathcal{F}_{\beta}\uparrow$  & $ \mathcal{M}\downarrow$\\
  \midrule
  1&Baseline (B)  &.830&.069   &.732&.056  &.736&.091 \\
  2&B + DSAM[+]  &.875&.055   &.771&.045  &.747&.077  \\
  3&B + DSAM[c]  &.873&.056    &.790&.043  &.753&.071   \\
  4&B + DSAM[*]  &\textbf{.889}&\textbf{.051}    &\textbf{.813}&\textbf{.039} &\textbf{.810}&\textbf{.062} \\
  \bottomrule
 \end{tabu}
 \vspace{-18pt}
 \label{tab:DSAM}
\end{table}

\begin{table*}[t]
   \caption{\small Ablation study of each module in DSAF.}
   \label{tab:each_comp}
\centering
\begin{adjustbox}{max width=\textwidth}
\renewcommand{\arraystretch}{1.0}	\setlength\aboverulesep{0.5pt}\setlength\belowrulesep{1pt}
   \setlength{\tabcolsep}{2.5mm}{
        \footnotesize
        \begin{tabu} to 0.99\textwidth{X[0.5,c] | X[6,c]| X[c] X[c] | X[c] X[c] | X[c] X[c] | X[c] X[c] | X[c] X[c] | X[c] X[c] | X[c] X[c]}

         \toprule
         \multicolumn{1}{c|}{\multirow{2}{*}{{\#}}}&
         \multicolumn{1}{c|}{\multirow{2}{*}{Settings}} & \multicolumn{2}{c|}{DUT-RGBD~} & \multicolumn{2}{c|}{NJUD~} & \multicolumn{2}{c|}{NLPR~}& \multicolumn{2}{c|}{SSD~}& \multicolumn{2}{c|}{STERE}& \multicolumn{2}{c|}{LFSD~}&
         \multicolumn{2}{c}{RGBD135~} \\
         && 
            $\mathcal{F}_{\beta}\uparrow$  & $ \mathcal{M}\downarrow$&
            $\mathcal{F}_{\beta}\uparrow$  & $ \mathcal{M}\downarrow$&
            $\mathcal{F}_{\beta}\uparrow$  & $ \mathcal{M}\downarrow$&
            $\mathcal{F}_{\beta}\uparrow$  & $ \mathcal{M}\downarrow$&
            $\mathcal{F}_{\beta}\uparrow$  & $ \mathcal{M}\downarrow$&
            $\mathcal{F}_{\beta}\uparrow$  & $ \mathcal{M}\downarrow$&
            $\mathcal{F}_{\beta}\uparrow$  & $ \mathcal{M}\downarrow$\\
         \midrule
         1&Baseline (B)  &.830&.069  &.821&.066  &.732 & .056  &.736 &.091  &.786 &.073  &.801&.092 &.762&.047\\
         2&B + DSAM    &.889&.051  &.853&.055  &.813 & .039  &.810 &.062  &.816 &.064  &.823&.083 &.823&.035\\
         3 & B + DSAM + ACMF
         &\textbf{.926}&\textbf{.030}  
         &\textbf{.901}&\textbf{.039} 
         &\textbf{.897}&\textbf{.024} 
         &\textbf{.852}&\textbf{.045}  
         &\textbf{.898}&\textbf{.036} 
         &\textbf{.882}&\textbf{.054}  
         &\textbf{.896}&\textbf{.021}  \\
         \bottomrule
   \end{tabu}}
    \end{adjustbox}
   \vspace{-15pt}
\end{table*}

\begin{figure}[h]
\centering
\includegraphics[width=3.3in]{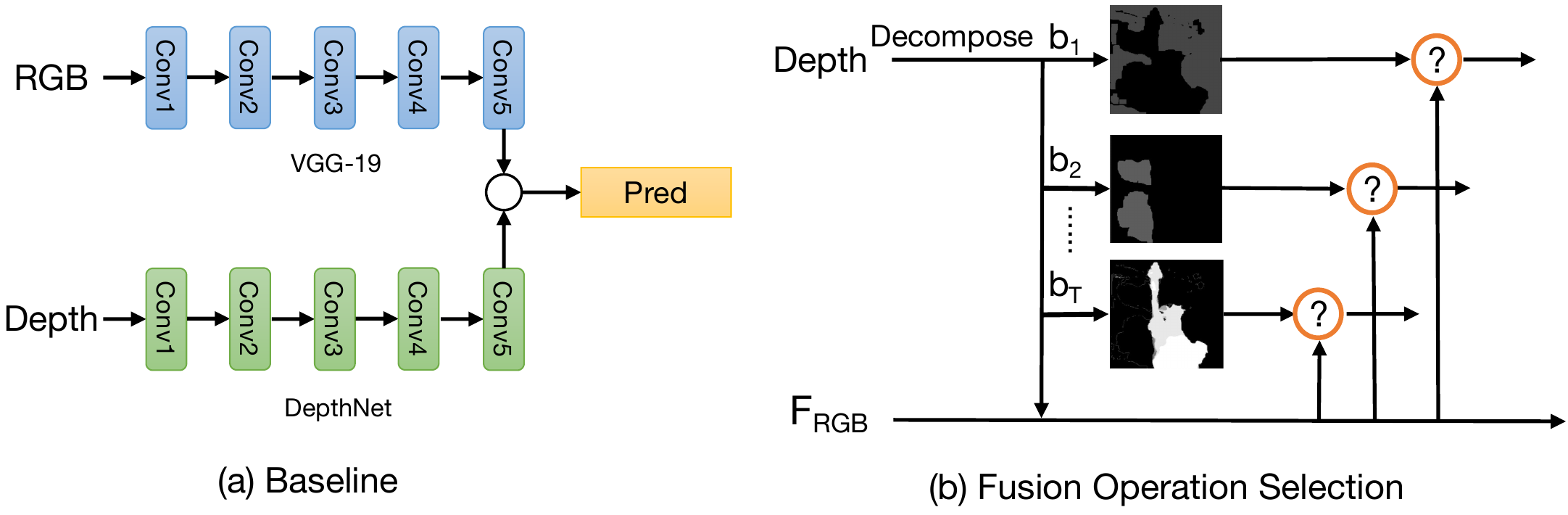}
\caption{The detailed illustration of our baseline is shown in (a), and the fusion operation selection is shown in (b). }
\label{fig:abadsam}
\vspace{-18pt}
\end{figure}

\begin{table}[h]
    \caption{\small Ablation analysis for the different region numbers $T+1$.}
    \label{tab:agg}
    \centering
    \footnotesize
    \begin{tabu} to 0.99\columnwidth{X[1.2,c]| X[c] X[c] | X[c] X[c] | X[c] X[c]}
     \toprule
     \multicolumn{1}{c|}
     {\multirow{2}{*}{$T+1$}} &
     \multicolumn{2}{c|}{DUT-RGBD~} &
     \multicolumn{2}{c|}{NLPR~}& 
     \multicolumn{2}{c}{SSD~} 
     \\
     &
     $\mathcal{F}_{\beta}\uparrow$  & $ \mathcal{M}\downarrow$&
     $\mathcal{F}_{\beta}\uparrow$  & $ \mathcal{M}\downarrow$&
     $\mathcal{F}_{\beta}\uparrow$  & $ \mathcal{M}\downarrow$\\
     \midrule
    1  &.854&.060  &.771&.047  &.755&.084  \\
    2  &.874&.056  &.781&.045  &.790& \textbf{.068} \\
    3  &\textbf{.889}&\textbf{.051}    &\textbf{.813}&\textbf{.039} &\textbf{.810}&.062\\
    4  &.844&.063   &.760&.049  &.743&.078   \\
    5 &.831 & .070   &.755&.048  &.715 & .098  \\
     \bottomrule
    \end{tabu}
    \label{tab:binnumber}
    \vspace{-10pt}
   \end{table}
   \vspace{-5pt}

In order to verify the effectiveness of the proposed depth-sensitive attention module, we conduct a series of experiments with different strategies: 
1) Baseline. The network contains a VGG-19 backbone for the RGB branch and a DepthNet for the depth branch, as shown in \cref{fig:abadsam} (a). 2-4) The network consists of the VGG-19 backbone equipped with different DSAMs and the DepthNet. As for the strategies 2-4, as shown in \cref{fig:abadsam} (b), we try different fusion operations of the depth masks and the RGB features for DSAM. The strategies 2,3,4 represent `element-wise summation' (+), `concatenation' (c), `element-wise multiplication' (*) operation, respectively. 
The results are shown in \cref{tab:DSAM}, and our DSAM can improve the baseline by a large margin, which demonstrates the effectiveness of DSAM. 
From \cref{tab:DSAM}, we observe that the `element-wise multiplication' operation obtains the best overall performance as it directly serves as the spatial attention mechanism. 
Moreover, the `concatenation' operation achieves the suboptimal accuracy, and we suspect that the depth cues play a role of `position encoding' here.

\vspace{-3pt}

\paragraph{Effect of the number of depth regions.} 
The region number of depth decomposition is an important hyper-parameter in our method, 
thus we perform the experiments with different $T+1$ values. Table \ref{tab:binnumber} lists the performance as $T$ varies, and DSAM achieves the best accuracy when $T+1$ is 3.

\begin{figure}[t]
   \centering
   \includegraphics[width=0.90\linewidth]{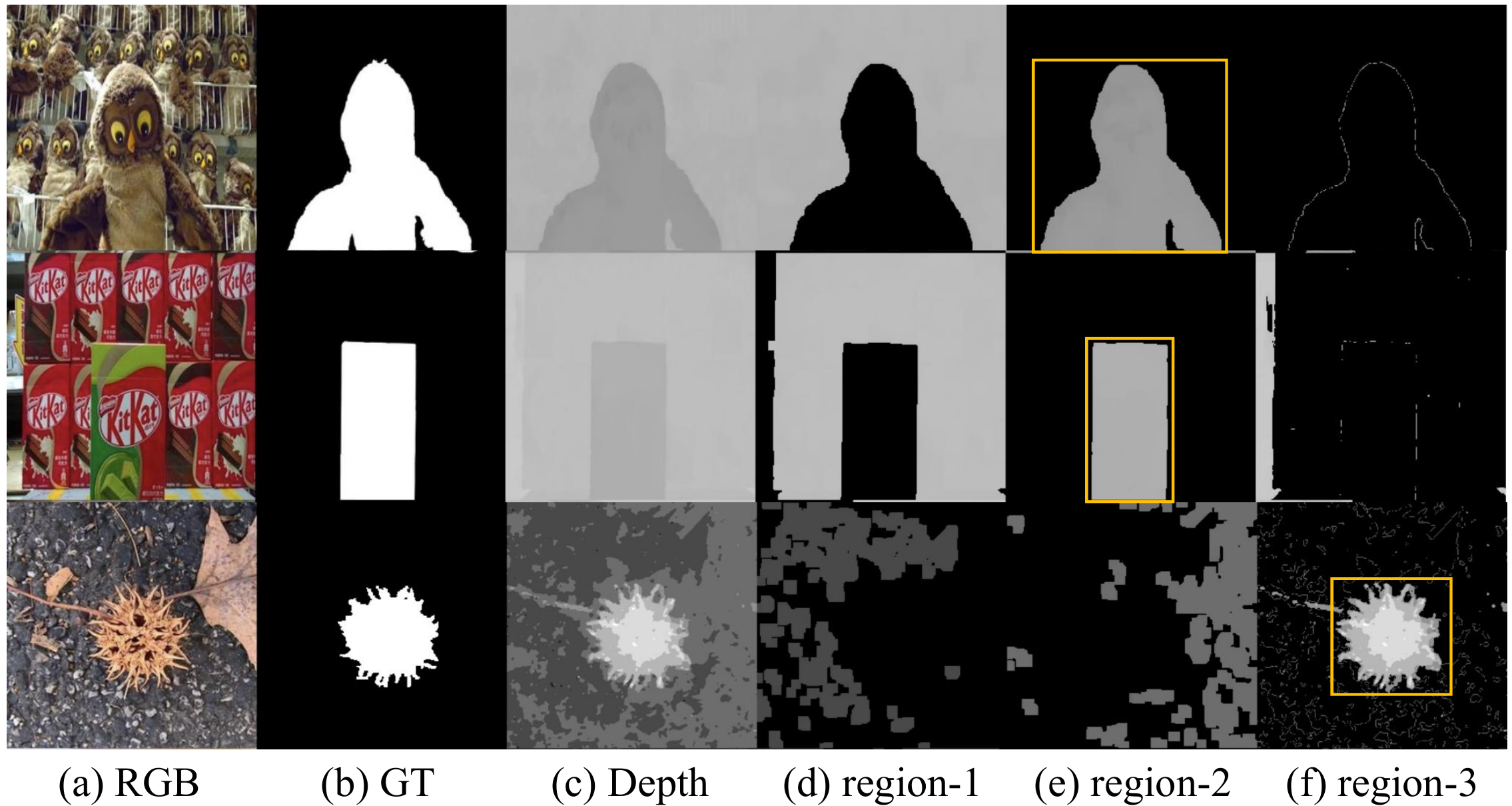}
   \caption{Visualization results of depth decomposition. Row (a), (b), (c) represent the RGB image, ground truth, depth map. Row (d), (e), (f) show the three regions of the depth decomposition.}
   \label{fig:split_result}
   \vspace{-18pt}
\end{figure}

\vspace{-3pt}

\paragraph{Effectiveness of the task-specific search space.}

In this part, we conduct the corresponding ablation studies to evaluate the  effectiveness of each type of cell in our multi-modal search space. We perform the architecture search process and retrain the whole network under different search spaces. The corresponding results are shown in \cref{search_space}. 

\vspace{-3pt}

\paragraph{Effectiveness of the attentions in our search space.}\label{para:ablation-atten}

To demonstrate the effectiveness of the attention operations, we perform the searching process with or without the spatial and channel attention operations. The corresponding results are shown in \cref{search_space}. With the injection of the attention operations, the performance of the model has a large improvement, which demonstrates that the attention mechanism plays an important role in RGB-D SOD. 

\begin{table}[t]
	\caption{Ablation study for the designed search space. MM, MS, GA, SR are four types of cells mentioned above. AT represents the attention operations in the search space.}
   \label{search_space}
   \footnotesize
   \begin{tabu} to 0.99\columnwidth{ X[0.6,c] X[0.6,c] X[0.6,c] X[0.6,c] X[0.6,c] | X[c] X[c] | X[c] X[c]}
      \toprule
      \multirow{2}{*}{MM} & \multirow{2}{*}{MS} & \multirow{2}{*}{GA} 
      & \multirow{2}{*}{SR} 
      & \multirow{2}{*}{AT} 
      & \multicolumn{2}{c|}{DUT-RGBD} & 
      \multicolumn{2}{c}{NLPR} \\
      &&&&&
      $\mathcal{F}_{\beta}\uparrow$ 
      & $ \mathcal{M}\downarrow$ & 
      $\mathcal{F}_{\beta}\uparrow$  
      & $ \mathcal{M}\downarrow$
      \\ 
      \midrule
      &   &   &  & & .889 & .051 & .813 & .039 \\
      \checkmark &&&&& .908 & .043 & .837 & .033 \\

      \checkmark &  \checkmark  &  & &   & .912 & .041 & .846 & .033 \\

      \checkmark & \checkmark  &  \checkmark  &  & & .918 & .037 & .857 & .030 \\

      \checkmark & \checkmark & \checkmark &  \checkmark & & .919 & .035 & .868 & .028 \\
      \midrule
      \checkmark &  \checkmark & \checkmark &  \checkmark & \checkmark & \textbf{.926} & \textbf{.030} & \textbf{.897} & \textbf{.024} \\
      \bottomrule
   \end{tabu}
   \vspace{-15pt}
\end{table}

\vspace{-3pt}

\paragraph{Searched architecture visualization.} Due to the limited space, we illustrate the searched fusion module in the supplementary material.
An interesting observation is that in the MM cell, the numbers of operations connected to RGB features are more than those connected to depth features. 
The phenomenon demonstrates that considering the differences between RGB and depth data, numerous redundant operations or channels of depth features are unnecessary, which also verifies the asymmetric two-stream architecture for RGB and depth branches in ATSA~\cite{Zhang_2020_ECCV} is reasonable.

\vspace{-3pt}

\paragraph{Effectiveness of each component in DSAF}

Table \ref{tab:each_comp} summarizes how performance
gets improved by adding each component step by step into our DSA$^2$F on seven standard benchmarks. The table shows that each component of our DSA$^2$F provides a significant performance gain.

\vspace{-3pt}

\section{Conclusion}

In this paper, we have proposed a two-stream framework named DSA$^2$F for RGB-D saliency detection.
In the framework, we have introduced a depth-sensitive attention module (DSAM) to effectively enhance the RGB features and reduce the background distraction by utilizing the depth geometry information.
Furthermore, we have designed a task-specific search space tailored for the multi-modal multi-scale feature fusion and obtained a powerful fusion architecture automatically.
Extensive experiments have demonstrated the effectiveness of our framework against previous state-of-the-art methods, and the visualization results have proved that our network is capable of precisely capturing salient regions in challenging scenes.

\footnotesize
\paragraph{Acknowledgements} This work is supported in part by National Key Research and Development Program of China under Grant 2020AAA0107400, National Natural Science Foundation of China under Grant U20A20222, Zhejiang Provincial Natural Science Foundation of China under Grant LR19F020004, and key scientific technological innovation research project by Ministry of Education.

{\small
\bibliographystyle{ieee_fullname}
\bibliography{egbib}

\begin{thebibliography}{10}\itemsep=-1pt

\bibitem{achanta2009frequency}
Radhakrishna Achanta, Sheila Hemami, Francisco Estrada, and Sabine Susstrunk.
\newblock Frequency-tuned salient region detection.
\newblock In {\em IEEE Conf. Comput. Vis. Pattern Recog.}, pages 1597--1604.
  IEEE, 2009.

\bibitem{Anandalingam1992HierarchicalOA}
G. Anandalingam and T. Friesz.
\newblock Hierarchical optimization: An introduction.
\newblock {\em Annals of Operations Research}, 34:1--11, 1992.

\bibitem{baker2017designing}
Bowen Baker, Otkrist Gupta, Nikhil Naik, and Ramesh Raskar.
\newblock Designing neural network architectures using reinforcement learning.
\newblock 2017.

\bibitem{Bender2018UnderstandingAS}
Gabriel Bender, P. Kindermans, Barret Zoph, V. Vasudevan, and Quoc~V. Le.
\newblock Understanding and simplifying one-shot architecture search.
\newblock In {\em Int. Conf. Mach. Learn.}, 2018.

\bibitem{borji2015salient}
Ali Borji, Ming-Ming Cheng, Huaizu Jiang, and Jia Li.
\newblock Salient object detection: A benchmark.
\newblock {\em IEEE Trans. Image Process.}, 24(12):5706--5722, 2015.

\bibitem{Brock2018SMASHOM}
A. Brock, T. Lim, J.~M. Ritchie, and N. Weston.
\newblock Smash: One-shot model architecture search through hypernetworks.
\newblock {\em Int. Conf. Learn. Represent.}, abs/1708.05344, 2018.

\bibitem{chen2018progressively}
Hao Chen and Youfu Li.
\newblock Progressively complementarity-aware fusion network for rgb-d salient
  object detection.
\newblock In {\em IEEE Conf. Comput. Vis. Pattern Recog.}, pages 3051--3060,
  2018.

\bibitem{Chen2019ThreeStreamAN}
H. Chen and Y. Li.
\newblock Three-stream attention-aware network for rgb-d salient object
  detection.
\newblock {\em IEEE Trans. Image Process.}, 28:2825--2835, 2019.

\bibitem{chenhaommf}
H. Chen, Y. Li, and Dan Su.
\newblock Multi-modal fusion network with multi-scale multi-path and
  cross-modal interactions for rgb-d salient object detection.
\newblock {\em Pattern Recognition}, 86:376--385, 2019.

\bibitem{Chen2019MultimodalFN}
H. Chen, Y. Li, and Dan Su.
\newblock Multi-modal fusion network with multi-scale multi-path and
  cross-modal interactions for rgb-d salient object detection.
\newblock {\em Pattern Recognition}, 86:376--385, 2019.

\bibitem{Chen2020DiscriminativeCT}
Hao Chen, Youfu Li, and Dan Su.
\newblock Discriminative cross-modal transfer learning and densely cross-level
  feedback fusion for rgb-d salient object detection.
\newblock {\em IEEE Transactions on Cybernetics}, 50:4808--4820, 2020.

\bibitem{chen2020eccv}
Shuhan Chen and Yun Fu.
\newblock Progressively guided alternate refinement network for rgb-d salient
  object detection.
\newblock In {\em Eur. Conf. Comput. Vis.}, 2020.

\bibitem{Chen2018ReinforcedEN}
Yukang Chen, Q. Zhang, C. Huang, Lisen Mu, Gaofeng Meng, and Xinggang Wang.
\newblock Reinforced evolutionary neural architecture search.
\newblock {\em ArXiv}, abs/1808.00193, 2018.

\bibitem{cheng2014depth}
Yupeng Cheng, Huazhu Fu, Xingxing Wei, Jiangjian Xiao, and Xiaochun Cao.
\newblock Depth enhanced saliency detection method.
\newblock In {\em Proceedings of international conference on internet
  multimedia computing and service}, pages 23--27, 2014.

\bibitem{ciptadi2013depth}
Arridhana Ciptadi, Tucker Hermans, and James~M Rehg.
\newblock An in depth view of saliency.
\newblock In {\em Brit. Mach. Vis. Conf.} Georgia Institute of Technology,
  2013.

\bibitem{Colson2007AnOO}
B. Colson, P. Marcotte, and G. Savard.
\newblock An overview of bilevel optimization.
\newblock {\em Annals of Operations Research}, 153:235--256, 2007.

\bibitem{fan2017structure}
Deng-Ping Fan, Ming-Ming Cheng, Yun Liu, Tao Li, and Ali Borji.
\newblock Structure-measure: A new way to evaluate foreground maps.
\newblock In {\em Int. Conf. Comput. Vis.}, pages 4548--4557, 2017.

\bibitem{fan2018enhanced}
Deng-Ping Fan, Cheng Gong, Yang Cao, Bo Ren, Ming-Ming Cheng, and Ali Borji.
\newblock Enhanced-alignment measure for binary foreground map evaluation.
\newblock {\em arXiv preprint arXiv:1805.10421}, 2018.

\bibitem{Fan2020RethinkingRS}
Deng-Ping Fan, Zheng Lin, Jiaxing Zhao, Y. Liu, Zhao Zhang, Q. Hou, Menglong
  Zhu, and Ming-Ming Cheng.
\newblock Rethinking rgb-d salient object detection: Models, datasets, and
  large-scale benchmarks.
\newblock {\em IEEE transactions on neural networks and learning systems}, PP,
  2020.

\bibitem{fan2019shifting}
Deng-Ping Fan, Wenguan Wang, Ming-Ming Cheng, and Jianbing Shen.
\newblock Shifting more attention to video salient object detection.
\newblock In {\em IEEE Conf. Comput. Vis. Pattern Recog.}, pages 8554--8564,
  2019.

\bibitem{fan2020bbsnet}
Deng-Ping Fan, Yingjie Zhai, Ali Borji, Jufeng Yang, and Ling Shao.
\newblock Bbs-net: Rgb-d salient object detection with a bifurcated backbone
  strategy network.
\newblock In {\em Eur. Conf. Comput. Vis.}, 2020.

\bibitem{fan2014salient}
Xingxing Fan, Zhi Liu, and Guangling Sun.
\newblock Salient region detection for stereoscopic images.
\newblock In {\em 2014 19th International Conference on Digital Signal
  Processing}, pages 454--458. IEEE, 2014.

\bibitem{feng2016local}
David Feng, Nick Barnes, Shaodi You, and Chris McCarthy.
\newblock Local background enclosure for rgb-d salient object detection.
\newblock In {\em IEEE Conf. Comput. Vis. Pattern Recog.}, pages 2343--2350,
  2016.

\bibitem{Fu2020JLDCF}
Keren Fu, Deng-Ping Fan, Ge-Peng Ji, and Qijun Zhao.
\newblock Jl-dcf: Joint learning and densely-cooperative fusion framework for
  rgb-d salient object detection.
\newblock In {\em IEEE Conf. Comput. Vis. Pattern Recog.}, pages 3052--3062,
  2020.

\bibitem{Gao20123DOR}
Yue Gao, M. Wang, D. Tao, R. Ji, and Q. Dai.
\newblock 3-d object retrieval and recognition with hypergraph analysis.
\newblock {\em IEEE Trans. Image Process.}, 21:4290--4303, 2012.

\bibitem{Ghiasi2019NASFPNLS}
G. Ghiasi, Tsung-Yi Lin, R. Pang, and Quoc~V. Le.
\newblock Nas-fpn: Learning scalable feature pyramid architecture for object
  detection.
\newblock {\em IEEE Conf. Comput. Vis. Pattern Recog.}, pages 7029--7038, 2019.

\bibitem{Hong2015OnlineTB}
Seunghoon Hong, Tackgeun You, Suha Kwak, and B. Han.
\newblock Online tracking by learning discriminative saliency map with
  convolutional neural network.
\newblock In {\em Int. Conf. Mach. Learn.}, 2015.

\bibitem{Wei_2020_ECCV}
Wei {Ji}, Jingjing {Li}, Miao {Zhang}, Yongri {Piao}, and Huchuan {Lu}.
\newblock Accurate rgb-d salient object detection via collaborative learning.
\newblock In {\em Eur. Conf. Comput. Vis.}, 2020.

\bibitem{ju2014depth}
Ran Ju, Ling Ge, Wenjing Geng, Tongwei Ren, and Gangshan Wu.
\newblock Depth saliency based on anisotropic center-surround difference.
\newblock In {\em IEEE Int. Conf. Image Process.}, pages 1115--1119. IEEE,
  2014.

\bibitem{Lang2012DepthMI}
C. Lang, T.~V. Nguyen, H. Katti, Karthik Yadati, M. Kankanhalli, and S. Yan.
\newblock Depth matters: Influence of depth cues on visual saliency.
\newblock In {\em Eur. Conf. Comput. Vis.}, 2012.

\bibitem{li2020}
Chongyi Li, Runmin Cong, Yongri Piao, Qianqian Xu, and Chen~Change Loy.
\newblock Rgb-d salient object eetection with cross-modality modulation and
  selection.
\newblock {\em Eur. Conf. Comput. Vis.}, 2020.

\bibitem{Li_2020_CMWNet}
Gongyang Li, Zhi Liu, Linwei Ye, Yang Wang, and Haibin Ling.
\newblock Cross-modal weighting network for rgb-d salient object detection.
\newblock In {\em Eur. Conf. Comput. Vis.}, 2020.

\bibitem{li2014saliency}
Nianyi Li, Jinwei Ye, Yu Ji, Haibin Ling, and Jingyi Yu.
\newblock Saliency detection on light field.
\newblock In {\em IEEE Conf. Comput. Vis. Pattern Recog.}, pages 2806--2813,
  2014.

\bibitem{Lin2020GraphGuidedAS}
Pei-Wen Lin, Peng Sun, G. Cheng, S. Xie, X. Li, and Jianping Shi.
\newblock Graph-guided architecture search for real-time semantic segmentation.
\newblock {\em IEEE Conf. Comput. Vis. Pattern Recog.}, pages 4202--4211, 2020.

\bibitem{autodeeplab2019}
Chenxi Liu, Liang-Chieh Chen, Florian Schroff, Hartwig Adam, Wei Hua, Alan
  Yuille, and Li Fei-Fei.
\newblock Auto-deeplab: Hierarchical neural architecture search for semantic
  image segmentation.
\newblock In {\em IEEE Conf. Comput. Vis. Pattern Recog.}, 2019.

\bibitem{Liu2013AMO}
Guanghai Liu and Deng-Ping Fan.
\newblock A model of visual attention for natural image retrieval.
\newblock {\em 2013 International Conference on Information Science and Cloud
  Computing Companion}, pages 728--733, 2013.

\bibitem{Liu2019DARTSDA}
Hanxiao Liu, K. Simonyan, and Yiming Yang.
\newblock Darts: Differentiable architecture search.
\newblock {\em Int. Conf. Learn. Represent.}, 2019.

\bibitem{Liu2020LearningSS}
Nian Liu, N. Zhang, and J. Han.
\newblock Learning selective self-mutual attention for rgb-d saliency
  detection.
\newblock {\em IEEE Conf. Comput. Vis. Pattern Recog.}, pages 13753--13762,
  2020.

\bibitem{Liu2019SalientOD}
Zhengyi Liu, S. Shi, Quntao Duan, W. Zhang, and Peng Zhao.
\newblock Salient object detection for rgb-d image by single stream recurrent
  convolution neural network.
\newblock {\em Neurocomputing}, 363:46--57, 2019.

\bibitem{Luo2020CascadeGN}
Ao Luo, Xin Li, Fan Yang, Zhicheng Jiao, Hong Cheng, and Siwei Lyu.
\newblock Cascade graph neural networks for rgb-d salient object detection.
\newblock In {\em Eur. Conf. Comput. Vis.}, 2020.

\bibitem{mahadevan2009saliency}
Vijay Mahadevan and Nuno Vasconcelos.
\newblock Saliency-based discriminant tracking.
\newblock In {\em IEEE Conf. Comput. Vis. Pattern Recog.}, pages 1007--1013.
  IEEE, 2009.

\bibitem{niu2012leveraging}
Yuzhen Niu, Yujie Geng, Xueqing Li, and Feng Liu.
\newblock Leveraging stereopsis for saliency analysis.
\newblock In {\em IEEE Conf. Comput. Vis. Pattern Recog.}, pages 454--461.
  IEEE, 2012.

\bibitem{HDFNet-ECCV2020}
Youwei Pang, Lihe Zhang, Xiaoqi Zhao, and Huchuan Lu.
\newblock Hierarchical dynamic filtering network for rgb-d salient object
  detection.
\newblock In {\em Eur. Conf. Comput. Vis.}, 2020.

\bibitem{Paszke2017AutomaticDI}
Adam Paszke, S. Gross, Francisco Massa, A. Lerer, J. Bradbury, G. Chanan, T.
  Killeen, Z. Lin, N. Gimelshein, L. Antiga, Alban Desmaison, Andreas K{\"o}pf,
  E. Yang, Zach DeVito, Martin Raison, Alykhan Tejani, Sasank Chilamkurthy, B.
  Steiner, Lu Fang, Junjie Bai, and Soumith Chintala.
\newblock Pytorch: An imperative style, high-performance deep learning library.
\newblock {\em Adv. Neural Inform. Process. Syst.}, 2019.

\bibitem{peng2014rgbd}
Houwen Peng, Bing Li, Weihua Xiong, Weiming Hu, and Rongrong Ji.
\newblock Rgbd salient object detection: a benchmark and algorithms.
\newblock In {\em Eur. Conf. Comput. Vis.}, pages 92--109. Springer, 2014.

\bibitem{PrezRa2019MFASMF}
Juan-Manuel P{\'e}rez-R{\'u}a, Valentin Vielzeuf, S. Pateux, M. Baccouche, and
  F. Jurie.
\newblock Mfas: Multimodal fusion architecture search.
\newblock {\em IEEE Conf. Comput. Vis. Pattern Recog.}, pages 6959--6968, 2019.

\bibitem{piao2019depth}
Yongri Piao, Wei Ji, Jingjing Li, Miao Zhang, and Huchuan Lu.
\newblock Depth-induced multi-scale recurrent attention network for saliency
  detection.
\newblock In {\em Int. Conf. Comput. Vis.}, pages 7254--7263, 2019.

\bibitem{Piao2020A2deleAA}
Yongri Piao, Zhengkun Rong, Miao Zhang, W. Ren, and Huchuan Lu.
\newblock A2dele: Adaptive and attentive depth distiller for efficient rgb-d
  salient object detection.
\newblock {\em IEEE Conf. Comput. Vis. Pattern Recog.}, pages 9057--9066, 2020.

\bibitem{qu2017rgbd}
Liangqiong Qu, Shengfeng He, Jiawei Zhang, Jiandong Tian, Yandong Tang, and
  Qingxiong Yang.
\newblock Rgbd salient object detection via deep fusion.
\newblock {\em IEEE Trans. Image Process.}, 26:2274--2285, 2017.

\bibitem{Real2019RegularizedEF}
E. Real, A. Aggarwal, Y. Huang, and Quoc~V. Le.
\newblock Regularized evolution for image classifier architecture search.
\newblock {\em ArXiv}, abs/1802.01548, 2019.

\bibitem{ren2015exploiting}
Jianqiang Ren, Xiaojin Gong, Lu Yu, Wenhui Zhou, and Michael Ying~Yang.
\newblock Exploiting global priors for rgb-d saliency detection.
\newblock In {\em IEEE Conf. Comput. Vis. Pattern Recog. Worksh.}, 2015.

\bibitem{Shigematsu2017LearningRS}
Riku Shigematsu, D. Feng, Shaodi You, and Nick Barnes.
\newblock Learning rgb-d salient object detection using background enclosure,
  depth contrast, and top-down features.
\newblock {\em 2017 IEEE International Conference on Computer Vision Workshops
  (ICCVW)}, pages 2749--2757, 2017.

\bibitem{Simonyan2015VeryDC}
K. Simonyan and Andrew Zisserman.
\newblock Very deep convolutional networks for large-scale image recognition.
\newblock {\em CoRR}, abs/1409.1556, 2015.

\bibitem{Song2017DepthAwareSO}
Hangke Song, Z. Liu, H. Du, Guangling Sun, Olivier~Le Meur, and T. Ren.
\newblock Depth-aware salient object detection and segmentation via multiscale
  discriminative saliency fusion and bootstrap learning.
\newblock {\em IEEE Trans. Image Process.}, 26:4204--4216, 2017.

\bibitem{Wang2015SaliencyawareGV}
Wenguan Wang, J. Shen, and F. Porikli.
\newblock Saliency-aware geodesic video object segmentation.
\newblock {\em IEEE Conf. Comput. Vis. Pattern Recog.}, pages 3395--3402, 2015.

\bibitem{Xu2019AutoFPNAN}
Hang Xu, Lewei Yao, Zhenguo Li, Xiaodan Liang, and W. Zhang.
\newblock Auto-fpn: Automatic network architecture adaptation for object
  detection beyond classification.
\newblock {\em Int. Conf. Comput. Vis.}, pages 6648--6657, 2019.

\bibitem{Yu2020DeepMN}
Zhou Yu, Yuhao Cui, Jun Yu, Meng Wang, Dacheng Tao, and Q. Tian.
\newblock Deep multimodal neural architecture search.
\newblock {\em ACM Int. Conf. Multimedia}, 2020.

\bibitem{Zhang2020UCNetUI}
Jing Zhang, Deng-Ping Fan, Yuchao Dai, S. Anwar, F. Saleh, T. Zhang, and N.
  Barnes.
\newblock Uc-net: Uncertainty inspired rgb-d saliency detection via conditional
  variational autoencoders.
\newblock {\em IEEE Conf. Comput. Vis. Pattern Recog.}, pages 8579--8588, 2020.

\bibitem{Zhang_2020_ECCV}
Miao Zhang, Sun~Xiao Fei, Jie Liu, Shuang Xu, Yongri Piao, and Huchuan Lu.
\newblock Asymmetric two-stream architecture for accurate rgb-d saliency
  detection.
\newblock In {\em Eur. Conf. Comput. Vis.}, 2020.

\bibitem{Zhang2020SelectSA}
Miao Zhang, W. Ren, Yongri Piao, Zhengkun Rong, and Huchuan Lu.
\newblock Select, supplement and focus for rgb-d saliency detection.
\newblock {\em IEEE Conf. Comput. Vis. Pattern Recog.}, pages 3469--3478, 2020.

\bibitem{zhao2019Contrast}
Jia-Xing Zhao, Yang Cao, Deng-Ping Fan, Ming-Ming Cheng, Xuan-Yi Li, and Le
  Zhang.
\newblock Contrast prior and fluid pyramid integration for rgbd salient object
  detection.
\newblock In {\em IEEE Conf. Comput. Vis. Pattern Recog.}, 2019.

\bibitem{zhao2013unsupervised}
Rui Zhao, Wanli Ouyang, and Xiaogang Wang.
\newblock Unsupervised salience learning for person re-identification.
\newblock In {\em IEEE Conf. Comput. Vis. Pattern Recog.}, pages 3586--3593,
  2013.

\bibitem{DANet}
Xiaoqi Zhao, Lihe Zhang, Youwei Pang, Huchuan Lu, and Lei Zhang.
\newblock A single stream network for robust and real-time rgb-d salient object
  detection.
\newblock In {\em Eur. Conf. Comput. Vis.}, 2020.

\bibitem{Zhu2019PDNetPG}
Chunbiao Zhu, X. Cai, Kan Huang, Thomas~H. Li, and G. Li.
\newblock Pdnet: Prior-model guided depth-enhanced network for salient object
  detection.
\newblock {\em Int. Conf. Multimedia and Expo}, pages 199--204, 2019.

\bibitem{zhu2017three}
Chunbiao Zhu and Ge Li.
\newblock A three-pathway psychobiological framework of salient object
  detection using stereoscopic technology.
\newblock In {\em IEEE Conf. Comput. Vis. Pattern Recog. Worksh.}, pages
  3008--3014, 2017.

\bibitem{DBLP:conf/iclr/ZophL17}
Barret Zoph and Quoc~V. Le.
\newblock Neural architecture search with reinforcement learning.
\newblock In {\em Int. Conf. Learn. Represent.}, 2017.

\end{thebibliography}
}

\end{document}